%% file: main.tex
\documentclass{article} %
\usepackage{kotex}
\usepackage{conference,times}

\input{math_commands.tex}

\usepackage{hyperref}
\usepackage{url}
\usepackage{epstopdf}
\usepackage{booktabs}
\usepackage{graphicx}
\usepackage[normalem]{ulem}

\usepackage{algorithm}
\usepackage{algpseudocode}

\title{SPACE-LoRA: Allocating Activation-Subspace Protection for Continual Learning}

\author{Seunghyun Yoo\thanks{Equal contribution.}\hspace{0.35em}, Kiseok Kim\footnotemark[1]\hspace{0.35em}, Hyeontae Joo\footnotemark[1]\hspace{0.35em}, Junyeop Bang \& Hwangnam Kim\thanks{Corresponding author.} \\
School of Electrical Engineering\\
Korea University\\
Seoul, Republic of Korea\\
\texttt{\{seunghyunyoo,kisuk528,motern800,junyeop9981,hnkim\}@korea.ac.kr}
}

\iclrfinalcopy
\begin{document}
\maketitle
\lhead{Preprint}
\begin{abstract}
This study addresses the catastrophic forgetting problem that occurs when sequentially learning successive tasks using Low-Rank Adaptation (LoRA) from a lifelong learning perspective. While existing approaches have primarily constrained parameter updates or learning subspaces to reduce interference with past knowledge, they have not fully considered additive interference. This occurs when a newly added residual adapter on top of a fixed past model generates non-zero responses along input directions important for old tasks, thereby altering previous predictions. To this end, we propose Subspace Protection with Allocated Capacity for Efficient Continual Adaptation (SPACE-LoRA). SPACE-LoRA directly suppresses the responses of the new residual branch along input activation directions that are important for old tasks and adaptively determines the protection coverage for each module based on past-task sensitivity estimated via a common Fisher sensitivity-based coverage target. Under a fixed LoRA rank, this approach adaptively adjusts module-specific protection coverage while suppressing interference along input directions sensitive to old tasks. We assess the effectiveness of activation-subspace protection in mitigating catastrophic forgetting and examine the role of sensitivity-guided protection in continual learning across diverse tasks. Code is available at https://anonymous.4open.science/r/SPACE-LoRA-7864.
\end{abstract}

\section{INTRODUCTION}
\label{sec:intro}

Continual learning (CL) requires a model to sequentially learn new tasks while retaining knowledge acquired from old tasks, demanding a balance between stability and plasticity~\citep{parisi2019continual}.
Parameter-efficient fine-tuning (PEFT), particularly Low-Rank Adaptation (LoRA), enables CL by learning low-rank updates while keeping the pretrained backbone frozen~\citep{hu2021lora}.
Existing LoRA-based methods address inter-task interference by separating the subspace for new task updates. O-LoRA uses an orthogonality regularization to separate current and previous LoRA subspaces~\citep{wang2023orthogonal}, whereas InfLoRA determines the LoRA input factor from the interference-free subspace orthogonal to previous tasks before learning each task and keeps the projection fixed during optimization~\citep{liang2024inflora}.
More recent methods further refine this approach. SplitLoRA determines the minor subspace dimension for each module through a first-order bound on aggregated old task gradients, balancing the old gradient energy falling into the subspace against the dimensionality available for new task learning~\citep{qiu2026splitlora}. JANUS-LoRA uses gradient rectification to decouple the LoRA factor updates, and reconstructs them through regularized least squares so that their composite weight update approximately follows the intended orthogonal direction~\citep{chen2026janus}.

However, freezing previously learned parameters does not guarantee that old task behavior is preserved. Even when these parameters remain unchanged, a newly learned LoRA branch can induce non-zero residual outputs on old task activations, which propagate to downstream representations and predictions.
Existing subspace-based methods reduce this interference by restricting new task updates relative to subspaces derived from old tasks.
However, in these methods, the new branch either still responds to old task activations or stops responding only because one LoRA factor is frozen.
Moreover, increasing protection reduces the subspace available for new task adaptation, creating a stability-plasticity trade-off in determining the protected subspace size.
Since historical sensitivity from the accumulated tasks is distributed differently across activation directions in different modules, a fixed protection size may protect some modules  not enough while unnecessarily constraining others. Moreover, sizing the protected subspace from aggregate old task gradients does not directly reflect this sensitivity associated with protected directions.

To address these limitations, we propose Subspace Protection with Allocated Capacity for Efficient Continual Adaptation (SPACE-LoRA), which directly constrains the input factor of each new task LoRA branch with respect to selected old task activation directions and independently selects the protection size of each module. For each module, SPACE-LoRA selects the minimum number of directions needed to reach the target coverage of past-task sensitivity.
Rather than using directions to determine the adaptation space for the new task, we directly constrain the LoRA input factor to the orthogonal complement of the protected subspace by projecting it after every optimizer step, while keeping both LoRA factors trainable.
As a result, the response of the newly added LoRA branch is identically zero for any activation within the protected subspace, regardless of the output factor. Meanwhile, the remaining orthogonal complement remains available for new task adaptation.
Because increasing the size of the protected subspace necessarily reduces the space available for new task adaptation, the protection capacity is chosen per module. Old task activations determine the candidate protection directions and their ordering within each module. Fisher information measures how much historical task sensitivity is captured as these directions are included. This coverage-based decision of SPACE-LoRA allows the protection dimensionality to vary according to the sensitivity distribution of each module. We evaluate SPACE-LoRA with a single coverage target across vision and language benchmarks under two training protocols, a wider range of setting than the methods we compare with. Mostly, SPACE-LoRA ranks first or close to first on every benchmark, whereas the strongest baseline changes from benchmark to benchmark.
\paragraph{Our main contributions are summarized as follows:}
\begin{itemize}
\item We formulate continual low-rank adaptation as a residual-response control problem and introduce a hard activation-subspace constraint that directly projects the input factor of each new LoRA branch. This constraint guarantees zero residual response on the selected historical activation subspace while leaving the complementary input space available for new-task adaptation.

\item We introduce a Fisher-coverage scheme that adapts the protection size to each module's sensitivity to previous tasks. By allocating stronger protection to modules that require greater preservation while avoiding unnecessary protection in less sensitive modules, SPACE-LoRA balances knowledge retention and adaptation capacity across the network.

\item We evaluate SPACE-LoRA on both vision and language models across a diverse set of continual learning benchmarks and against a broad range of competitive baselines. The results show that SPACE-LoRA consistently alleviates catastrophic forgetting while preserving adaptation to incoming tasks, thereby improving the trade-off between stability and plasticity across different model architectures and task domains.
\end{itemize}

\section{RELATED WORK}
\label{sec:rework}
Continual learning (CL) studies how models learn from sequential tasks while retaining previously acquired knowledge from earlier tasks, and the main difficulty is catastrophic forgetting~\citep{de2021continual}.
CL is categorized as task-incremental learning, where the task identity is given, class-incremental learning, where the model chooses among all classes seen, and domain-incremental learning, where the input distribution changes while the label space stays fixed~\citep{van2022three}.
Existing CL methods are commonly categorized into regularization-based, memory-based, and expansion-based approaches~\citep{wang2024comprehensive}. Parameter-efficient fine-tuning (PEFT), which freezes a pretrained model and trains only a small number of additional parameters, can reduce the computational cost of adapting to new downstream tasks while maintaining competitive performance. Representative PEFT approaches include adapter-based tuning, which augments a frozen backbone with smaller and trainable modules~\citep{houlsby2019parameter}, and prompt-based tuning, which optimizes learnable prompts while leaving the pretrained parameters fixed~\citep{lester2021power, wang2022learning}.
Low-Rank Adaptation (LoRA) keeps the pretrained model frozen and parametrizes updates to selected weight matrices with two low-rank factors~\citep{hu2021lora}. It has been increasingly adopted in CL methods that mitigate interference among sequential task updates.

O-LoRA reduces task interference by placing low-rank updates for different tasks in orthogonal parameter subspaces~\citep{wang2023orthogonal}. InfLoRA constructs the input factor from a subspace designed to avoid interference with old tasks and keeps the factor fixed while learning the new task~\citep{liang2024inflora}. PLAN pre-assigns orthogonal basis vectors to each task and selects those least sensitive to interference with previously learned parameters~\citep{wang2025plan}. LB-CL~\citep{qiao2024learn} uses the sensitivity of low-rank parameters to identify knowledge-specific components and incorporates them into gradient projection. SplitLoRA partitions the old task gradient space into principal and minor subspaces and adjusts the partition size across modules to balance stability and plasticity~\citep{qiu2026splitlora}. JANUS-LoRA targets zero change in the layer output on previous activations~\citep{chen2026janus}. It projects the full-weight update onto the complement of a historical activation subspace and, because independent factor updates break this projection. reconstruct the factor updates by regularized least squares so that their composite update approximately follows the projected direction.

These LoRA-based approaches are related to projection-based CL methods that restrict new updates with respect to subspaces associated with old task knowledge. GPM constructs important subspaces from old task representations, determining the size based on a layer-wise energy threshold, and projects new gradients onto their orthogonal complements~\citep{saha2021gradient}. Energy-based rank selection has also shown its effectiveness in LoRA-based CL~\citep{li2026energy}. Adam-NSCL estimates a null space from old task input-feature covariance and constrains subsequent optimization to that space by projecting the optimizer update~\citep{wang2021training}. Regularization-based methods such as EWC~\citep{kirkpatrick2017overcoming} quantify parameter importance for old tasks using the Fisher information and restrict changes to parameters that are important for retaining old task knowledge. A separate line of work, such as AdaLoRA and OA-Adapter, allocate adaptation capacity selectively across components to improve parameter efficiency~\citep{zhang2023adalora, wan2025adaptive}.

In subspace-based methods such as GPM and SplitLoRA, the protection size is determined by representation energy or averaged old task gradients, neither of which measures the per-example loss sensitivity along the activation directions that are actually protected.
 SPACE-LoRA treats selected old-task activation directions as a protected subspace and projects the input factor onto its orthogonal complement after each optimizer step, while keeping both LoRA factors trainable. As a result, the newly added LoRA branch produces zero response to activations within the protected subspace. In each module, historical activation directions are ordered by decreasing activation eigenvalue, and the protection size is selected as the minimum number of directions required to reach the target Fisher coverage.

\section{PRELIMINARIES}
\label{sec:prelim}

\subsection{LoRA-based Continual Learning}
Consider a task stream
\(\mathcal{T} = \{t_1, \dots, t_T\}\).
The objective is to learn new tasks while retaining
performance on previously learned tasks~\citep{zhou2024continual}.
We consider a setting in which task-specific LoRA adapters
are accumulated~\citep{wang2023orthogonal}.
After learning task \(t\), the effective weight of module \(m\)
combines its pre-trained weight
\(W_m^0 \in \mathbb{R}^{d_{\mathrm{out},m} \times d_{\mathrm{in},m}}\)
with the accumulated adapters:
\begin{equation}
W_m^{(t)} = W_m^0 + \eta \sum_{\tau=1}^{t} B_{m,\tau}A_{m,\tau}, \qquad \eta = \frac{\alpha}{r},
\end{equation}
where \(r\) denotes the LoRA rank, \(\alpha\) is the scaling
hyperparameter~\citep{hu2021lora}, and
\begin{equation}
A_{m,t} \in \mathbb{R}^{r \times d_{\mathrm{in},m}}, \qquad B_{m,t} \in \mathbb{R}^{d_{\mathrm{out},m} \times r}.
\end{equation}
During training on task \(t\), only \(A_{m,t}\) and \(B_{m,t}\)
within module \(m\) are updated, while \(W_m^0\) and all
adapters from previous tasks remain frozen. 

\subsection{Adapter Interference and Catastrophic Forgetting}
Even with previously learned adapters frozen,
a new adapter can change the output of a module.
For a fixed input activation
\(x \in \mathbb{R}^{d_{\mathrm{in},m}}\)
from a previous task, the output change induced by
the new adapter at module \(m\) is
\begin{equation}
\Delta h_m^{(t)}(x) = \eta B_{m,t}A_{m,t}x.
\end{equation}
This expression isolates the new adapter's contribution
at a fixed module input. Changes in preceding modules
may additionally alter the input activation itself.
Because each new adapter is optimized for the current task,
its contribution may disrupt predictions on previous tasks.
Such changes can lead to catastrophic forgetting.
Let \(a_{j,j}\) denote the accuracy of task \(j\)
immediately after learning it, and let \(a_{t,j}\)
denote its accuracy after learning through task
\(t\;(t>j)\).
Task-wise forgetting is defined as
\begin{equation}
F_j^{(t)} = a_{j,j} - a_{t,j}.
\end{equation}
Thus, \(F_j^{(t)}>0\) indicates a decline in accuracy
on task \(j\) following subsequent adaptation.

\begin{figure*}[t]
    \centering
    \includegraphics[
        width=0.96\textwidth,
        height=0.21\textheight,
        keepaspectratio
    ]{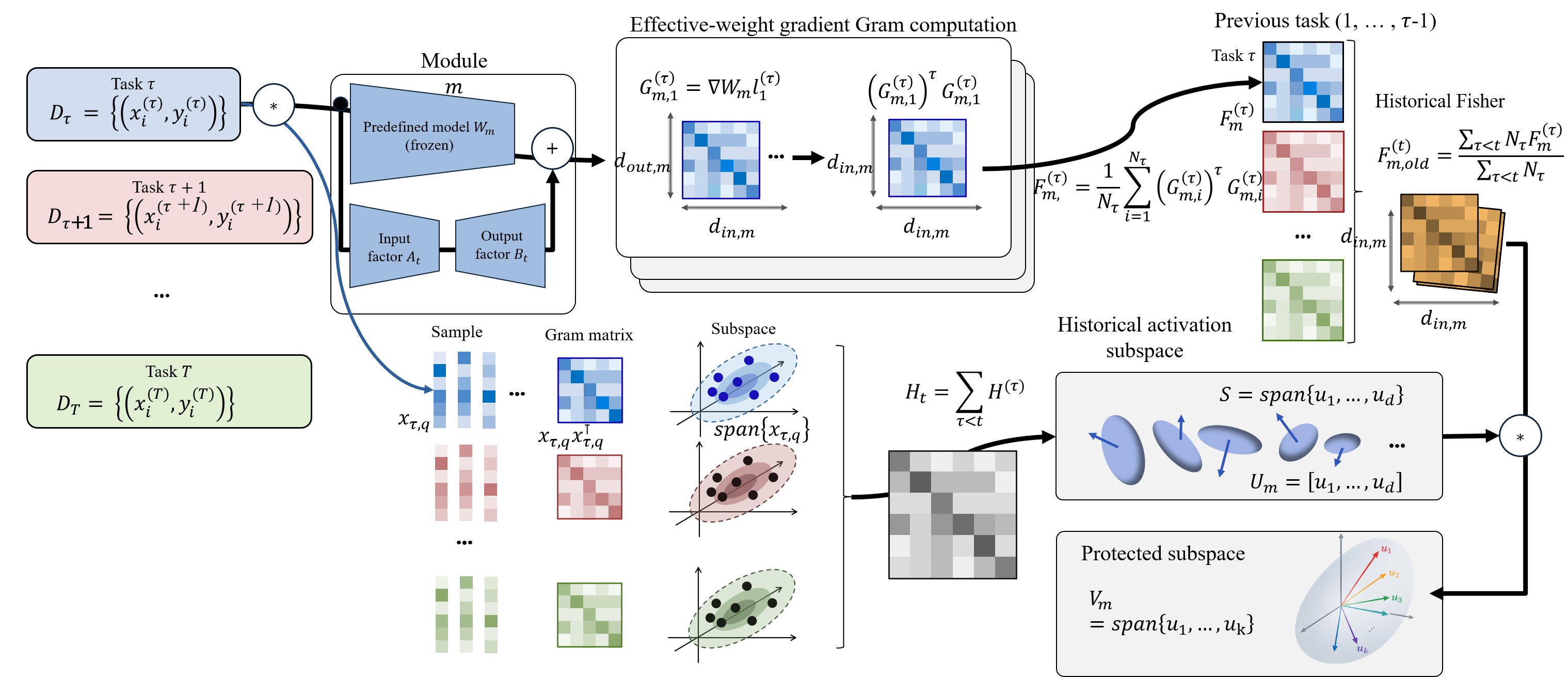}
    \caption{Overview of SPACE-LoRA.}
    \label{fig:spacelora_arch}
\end{figure*}

\section{SPACE-LoRA}
\label{sec:main}
We propose SPACE-LoRA as shown in Figure~\ref{fig:spacelora_arch}.
SPACE-LoRA directly suppresses the response of a newly added residual adapter along selected past activation directions, while determining the protection size of each module using a common Fisher sensitivity coverage target.
Under a fixed LoRA rank, this design suppresses interference along input directions that are sensitive to previous tasks while leaving the complementary input subspace available for learning the new task, thereby aiming to achieve a better trade-off between knowledge retention and adaptation.
In the subsequent sections, we describe how to construct the protected activation subspace and how to determine the required protection size for each module.

\subsection{Residual Response Nullification}
\label{sec:orthgonal}
As discussed in Section~\ref{sec:prelim}, even when all previously learned adapters are frozen, a newly added residual branch can alter past predictions if it produces a non-zero response to past-task inputs.
Therefore, continual adaptation should protect not only previously learned parameters but also the responses of the new adapter to input directions associated with old tasks.
Recent LoRA-based continual learning methods address this issue by constraining new-task adaptation with respect to subspaces associated with previously learned tasks.
InfLoRA constructs a low-interference adaptation subspace by removing current activation components aligned with historical gradient subspaces, whereas Janus-LoRA constrains the effective weight update to produce no change on historical activations.
Motivated by these approaches, we instead directly constrain the response of the newly introduced residual branch by making its input factor insensitive to selected historical activation directions.

At the endpoint of each task \(\tau\), we collect the input activations of each linear module \(m\) over the task training set \(D_\tau\) using the model obtained after completing the task.
Before learning task \(t\), these activations are summarized by the Historical Gram matrix
\begin{equation}
H_m^{<t} = \sum_{\tau<t} \sum_{x\in D_\tau} \sum_q x_{m,q}x_{m,q}^{\top},
\end{equation}
where \(x_{m,q}\) denotes the activation vector at the \(q\)-th input position of module \(m\).
We decompose
\begin{equation}
H_m^{<t} = U_m\Lambda_mU_m^\top,
\end{equation}
where the eigenvectors in \(U_m=[u_{m,1},\ldots,u_{m,d_m}]\) are ordered by decreasing eigenvalue.
These eigenvectors identify the principal input directions occupied by past activations.
Protecting the full basis would leave no input direction available to the new adapter.
We therefore protect only the first \(k_m\) activation directions and define
\begin{equation}
V_m = [u_{m,1},\ldots,u_{m,k_m}], \qquad P_m = V_mV_m^\top.
\end{equation}
Let \(A_m\) denote the input factor of the new LoRA branch.
After each optimizer update, we apply the hard projection
\begin{equation}
A_m \leftarrow A_m(I-P_m),
\label{eq:proj}
\end{equation}
which maintains
\begin{equation}
A_mV_m=0.
\end{equation}
Hence, for any \(x\in\operatorname{span}(V_m)\) with \(x=V_mc\),
\begin{equation}
\Delta W_m x = B_mA_mV_mc = 0.
\end{equation}
Thus, the newly added branch produces no residual response along the protected historical directions, while the orthogonal complement remains available for learning the new task.
This is a hard parameter constraint rather than a soft orthogonality regularizer.
A detailed characterization of its exact nullification property and its behavior outside the protected subspace is provided in Appendix~\ref{app:formal_properties}.

\subsection{Module-wise Fisher-Coverage Protection}
\label{sec:fisher}
The number of protected activation directions need not be identical across modules, since their importance to previously learned tasks can differ substantially.
Using a fixed protection size may therefore either leave sensitive directions insufficiently protected or unnecessarily reduce the input space available for new-task adaptation.
We instead determine \(k_m\) independently for each module as the minimum number of activation-ordered directions required to capture a prescribed fraction of its historical loss sensitivity.

To quantify this sensitivity, after completing task \(\tau\), we compute the gradient of each training example with respect to the effective weight of module \(m\),
\begin{equation}
G_{m,i}^{(\tau)}
=
\nabla_{W_m}\ell_i^{(\tau)}
\in
\mathbb{R}^{d_{\mathrm{out},m}\times d_{\mathrm{in},m}},
\end{equation}
and define the input-space empirical Fisher
\begin{equation}
F_m^{(\tau)}
=
\frac{1}{N_\tau}
\sum_{i=1}^{N_\tau}
\left(G_{m,i}^{(\tau)}\right)^\top
G_{m,i}^{(\tau)}.
\end{equation}
This statistic is computed at the endpoint of task \(\tau\) using the current task classifier and the observed ground-truth labels.
The historical Fisher available before task \(t\) is
\begin{equation}
F_{m,\mathrm{past}}^{(t)}
=
\frac{
\sum_{\tau<t}N_\tau F_m^{(\tau)}
}{
\sum_{\tau<t}N_\tau
}.
\end{equation}
Further details on the construction and accumulated historical Fisher are provided in Appendix~\ref{app:historical_statistics}.

Let
\(
U_m=[u_{m,1},\ldots,u_{m,d_m}]
\)
be the full activation basis from the previous subsection, ordered by decreasing activation eigenvalue, where \(d_m=d_{\mathrm{in},m}\).
We measure the historical sensitivity of each activation direction as
\begin{equation}
p_{m,j}^{(t)}
=
u_{m,j}^{\top}
F_{m,\mathrm{past}}^{(t)}
u_{m,j}.
\end{equation}
The activation statistics therefore determine the directions and their order, while the Fisher evaluates the loss sensitivity associated with those directions.
For a prefix of length \(k\), we define its Fisher coverage as
\begin{equation}
C_m(k)
=
\frac{
\sum_{j=1}^{k}p_{m,j}^{(t)}
}{
\operatorname{tr}
\left(
F_{m,\mathrm{past}}^{(t)}
\right)
}.
\end{equation}
Since \(U_m\) is a complete orthonormal basis,
\(
\operatorname{tr}(F_{m,\mathrm{past}}^{(t)})
=
\sum_{j=1}^{d_m}p_{m,j}^{(t)}
\),
so \(C_m(k)\) represents the fraction of total historical sensitivity captured by the first \(k\) activation directions.
Given a common coverage target \(\rho\), we select
\begin{equation}
k_m
=
\min
\left\{
k\in\{0,\ldots,d_m-r\}
:
C_m(k)\ge\rho
\right\}.
\end{equation}
The upper bound \(d_m-r\) leaves at least \(r\) input dimensions available to the rank-\(r\) adapter.
If the total sensitivity is zero, we set \(k_m=0\). If no admissible prefix reaches the coverage target, we set \(k_m=d_m-r\), corresponding to the maximum allowable protection size.
The resulting first \(k_m\) activation directions are protected using the hard response constraint in Section~\ref{sec:orthgonal}, and the protected subspace remains fixed throughout the current task.

\subsection{Update Procedure}
\label{sec:overall_procedure}
Algorithm~\ref{alg:space_lora_overview} summarizes the overall training procedure of SPACE-LoRA.
Before learning a new task, each module constructs its protected activation basis from the historical activation statistic and determines the protection size using the Fisher-coverage rule in Section~\ref{sec:fisher}.
The resulting protected subspace is then fixed throughout the current task.
A new LoRA branch is trained while its input factor is projected onto the orthogonal complement of the protected subspace after each optimizer update, as described in Section~\ref{sec:orthgonal}.
After completing the task, the endpoint activation and Fisher statistics are computed and incorporated into the historical statistics used for subsequent tasks.
The first task is trained without protection because no historical statistics are available.
Algorithm~\ref{alg:space_lora} in the appendix provides the complete procedure, including the construction and incremental update of the historical statistics.

\begin{algorithm}[!t]
\caption{Overview of SPACE-LoRA}
\label{alg:space_lora_overview}
\begin{algorithmic}[1]
\Require Task datasets $\{D_t\}_{t=1}^{T}$, LoRA rank $r$, coverage target $\rho$
\For{$t=1,\ldots,T$}
    \State Freeze the base model and all previously learned adapters
    \State Add a new rank-$r$ LoRA branch for task $t$
    
    \For{each protected module $m$}
        \If{$t=1$}
            \State Set $k_m\gets 0$ and $P_m\gets 0$
        \Else
            \State Obtain the activation-ordered basis from $H_m^{<t}$
            \State Determine $k_m$ using the Fisher-coverage target $\rho$
            \State Construct $V_m=[u_{m,1},\ldots,u_{m,k_m}]$ and $P_m=V_mV_m^\top$
        \EndIf
    \EndFor
    
    \For{each optimization step on $D_t$}
        \State Update the new LoRA branch using the current-task loss
        \State Project $A_{m,t}\gets A_{m,t}(I-P_m)$ for each protected module $m$
    \EndFor
    
    \State Update the historical activation and Fisher statistics using $D_t$
\EndFor
\end{algorithmic}
\end{algorithm}

\section{EMPIRICAL STUDY}

\subsection{Experimental Setup}

\paragraph{Datasets and Baselines.}
We evaluate SPACE-LoRA on vision and language continual learning benchmarks.
Our vision benchmarks comprise ImageNet-R, DomainNet, VTAB, and VDD.
For language continual learning, we use a suite of 15 NLP classification datasets following the benchmark used by Progressive Prompts and O-LoRA~\citep{razdaibiedina2023progressive,wang2023orthogonal}.
We conduct systematic comparisons with representative LoRA-based
continual learning methods: InfLoRA, SplitLoRA, and Janus-LoRA
for vision, and these methods together with O-LoRA for language.
The comparisons use a common experimental protocol within each
benchmark.
For methods originally developed for vision, their language
experiments use T5 adaptations based on the official implementations. The adaptation details are provided in the appendix.

\paragraph{Experimental Design.}
We evaluate ImageNet-R, DomainNet, and VTAB under
class-incremental learning (CIL), where prediction is performed
over all classes observed so far without access to the task
identity at inference.
For ImageNet-R, we follow the task-count configurations used by
\citet{qiu2026splitlora} and partition its 200 classes into
5, 10, and 20 tasks, with 40, 20, and 10 classes per task,
respectively.
For DomainNet, we use five domains excluding Quickdraw and
partition its 345 classes into five tasks of 69 classes each.
For VTAB, we adopt the 5-task cross-domain construction
introduced by APER, with 10 classes per task.
For VDD, we use a task-incremental learning (TIL) setting:
ImageNet is excluded, and each of the remaining nine datasets
constitutes a separate task, with task identity available
at inference.
For language continual learning, each dataset constitutes
one task.
We consider sequences of 5, 10, and 15 tasks, constructed
from the corresponding prefixes of a fixed 15-dataset order.
Shared tasks use the same data examples across these configurations.
Evaluation uses the candidate label set of the dataset being
evaluated, rather than a joint label space spanning all datasets.
We use pretrained ViT-B/16 and T5-large backbones for the
vision and language experiments, respectively.
Detailed dataset statistics, data splits, task sequences,
and implementation settings are provided in
Appendix~\ref{sec:appendix_exp}.

\paragraph{Evaluation Metrics and Implementation Details.}
We evaluate performance using average accuracy,
$\mathrm{ACC}_{t}=\frac{1}{t}\sum_{j=1}^{t}a_{t,j}$.
Class-incremental evaluation considers all seen classes
without task identity, whereas task-incremental evaluation
uses task identity to select the corresponding head.
For vision experiments, we use ViT-B/16 pretrained on
ImageNet-21K, with rank-$10$ LoRA adapters applied to the
key and value projections in all 12 transformer blocks.
All vision methods follow the same task splits and training
protocol within each benchmark.
We train for 50 epochs per task on ImageNet-R and VTAB,
and 5 epochs on DomainNet.
For VDD, we train for 1,500 optimization steps per task.
For comparisons with the baselines, we report SPACE-LoRA
with Fisher coverage thresholds of $\rho=0.90$ and $0.95$.
For language experiments, we use T5-large as the primary
backbone and follow the training settings of O-LoRA,
including the number of training epochs, batch size,
and learning rate.

\begin{table}[t]
\centering
\caption{
Class-incremental learning results on ImageNet-R.
}
\label{tab:imagenetr_main}
\small
\setlength{\tabcolsep}{4pt}
\resizebox{\linewidth}{!}{
\begin{tabular}{lcccccc}
\toprule
& \multicolumn{6}{c}{ImageNet-R} \\
\cmidrule(lr){2-7}
& \multicolumn{2}{c}{5 Tasks}
& \multicolumn{2}{c}{10 Tasks}
& \multicolumn{2}{c}{20 Tasks} \\
\cmidrule(lr){2-3}\cmidrule(lr){4-5}
\cmidrule(lr){6-7}
Method
& Acc. $\uparrow$ & F $\downarrow$
& Acc. $\uparrow$ & F $\downarrow$
& Acc. $\uparrow$ & F $\downarrow$ \\
\midrule
Joint-LoRA
& 82.68 & -
& 82.68 & -
& 82.68 & - \\
\midrule
InfLoRA
& $78.89 \pm 0.55$ & 6.02
& $75.90 \pm 0.22$ & 6.28
& $\mathbf{72.23 \pm 0.58}$ & 6.82 \\
SplitLoRA
& $74.46 \pm 0.22$ & 10.80
& $70.25 \pm 0.32$ & 7.98
& $59.84 \pm 0.25$ & \textbf{4.81} \\
Janus-LoRA
& $\mathbf{79.47 \pm 0.47}$ & 6.41
& $\mathbf{76.07 \pm 0.45}$ & 7.70
& $72.15 \pm 0.36$ & 8.22 \\
SPACE-LoRA ($\rho=0.90$)
& $79.09 \pm 0.31$ & \textbf{4.82}
& $74.64 \pm 0.44$ & 6.42
& $69.65 \pm 0.50$ & 6.80 \\
SPACE-LoRA ($\rho=0.95$)
& $78.98 \pm 0.33$ & 4.90
& $74.33 \pm 0.51$ & \textbf{5.98}
& $69.33 \pm 0.31$ & 5.34 \\
\bottomrule
\end{tabular}
}
\end{table}

\subsection{Main Results}

\paragraph{Results on ImageNet-R.}
We first evaluate SPACE-LoRA on ImageNet-R,
where successive tasks introduce disjoint subsets
of its classes.
Table~\ref{tab:imagenetr_main} reports the results
for three task partitions of ImageNet-R.
On ImageNet-R, both SPACE-LoRA configurations achieve
slightly higher mean final accuracy and lower forgetting
than InfLoRA in the 5-task setting, with $\rho=0.90$
yielding the lowest forgetting among continual learning methods.
In the 10-task setting, $\rho=0.95$ achieves the lowest
forgetting, whereas $\rho=0.90$ exhibits slightly more
forgetting than InfLoRA.
However, both configurations achieve lower final accuracy
than InfLoRA and Janus-LoRA in the 10- and 20-task settings.
These results highlight the importance of balancing
old task protection with new task adaptation.

\begin{table}[t]
\centering
\caption{
Continual learning performance on DomainNet, VTAB, and VDD.
DomainNet and VTAB use class-incremental evaluation,
whereas VDD uses task-incremental evaluation.
}
\label{tab:diverse_vision_main}
\small
\setlength{\tabcolsep}{4pt}
\begin{tabular}{lcccccc}
\toprule
& \multicolumn{2}{c}{DomainNet: 5 Tasks}
& \multicolumn{2}{c}{VTAB: 5 Tasks}
& \multicolumn{2}{c}{VDD: 9 Tasks} \\
\cmidrule(lr){2-3}\cmidrule(lr){4-5}\cmidrule(lr){6-7}
Method
& Acc. $\uparrow$ & F $\downarrow$
& Acc. $\uparrow$ & F $\downarrow$
& Acc. $\uparrow$ & F $\downarrow$ \\
\midrule
InfLoRA
& 72.44 & 7.81
& 85.52 & 6.74
& 80.88 & \textbf{1.86} \\
SplitLoRA
& 67.51 & 9.81
& 73.35 & 19.67
& 64.78 & 3.73 \\
Janus-LoRA
& \textbf{73.41} & 7.82
& 80.47 & 10.59
& 78.53 & 4.16 \\
SPACE-LoRA ($\rho=0.90$)
& 73.26 & 6.06
& \textbf{86.05} & \textbf{5.94}
& \textbf{86.26} & 3.32 \\
SPACE-LoRA ($\rho=0.95$)
& 73.19 & \textbf{5.54}
& 85.20 & 6.51
& 80.41 & 2.66 \\
\bottomrule
\end{tabular}
\end{table}

\paragraph{Results on DomainNet, VTAB, and VDD.}
We consider benchmarks encompassing multiple
visual domains or distinct classification tasks.
Table~\ref{tab:diverse_vision_main} shows that
SPACE-LoRA achieves the lowest forgetting on DomainNet
with $\rho=0.95$, while achieving slightly lower final accuracy than Janus-LoRA.
On VTAB, SPACE-LoRA with $\rho=0.90$ achieves the highest accuracy
of 86.05\% and the lowest forgetting of
5.94 percentage points.
On VDD, SPACE-LoRA reaches 86.26\% accuracy with
$\rho=0.90$, outperforming other baselines.
Both coverage settings outperform Janus-LoRA and
SplitLoRA in accuracy on VTAB and VDD.
These results show that SPACE-LoRA achieves competitive
performance across benchmarks involving multiple visual
domains or heterogeneous task sequences.

\begin{table}[t]
\centering
\caption{
Continual learning performance on the 5-, 10-, and 15-task
language benchmarks.}
\label{tab:language_cl}
\small
\setlength{\tabcolsep}{5pt}
\begin{tabular}{lcccccc}
\toprule
& \multicolumn{2}{c}{5 Tasks}
& \multicolumn{2}{c}{10 Tasks}
& \multicolumn{2}{c}{15 Tasks} \\
\cmidrule(lr){2-3}
\cmidrule(lr){4-5}
\cmidrule(lr){6-7}
Method
& Acc. $\uparrow$ & F $\downarrow$
& Acc. $\uparrow$ & F $\downarrow$
& Acc. $\uparrow$ & F $\downarrow$ \\
\midrule
O-LoRA
& 48.76 & 27.70
& 54.56 & 22.21
& 37.52 & 32.48 \\
InfLoRA
& 71.68 & \textbf{-0.90}
& 70.83 & \textbf{0.48}
& 67.54& 3.91 \\
SplitLoRA
& 61.60 & 0.40
& 58.72 & 0.54
& 54.85 & \textbf{0.50} \\
Janus-LoRA
& 70.96 & -0.75
& 66.97 & 2.96
& 63.08 & 2.26 \\
SPACE-LoRA ($\rho=0.90$)
& \textbf{71.88} & -0.10
& \textbf{73.95} & 0.98
& \textbf{69.79}  & 2.53 \\
\bottomrule
\end{tabular}
\end{table}

\begin{figure}[!ht]
    \centering
    \includegraphics[width=0.7\linewidth]{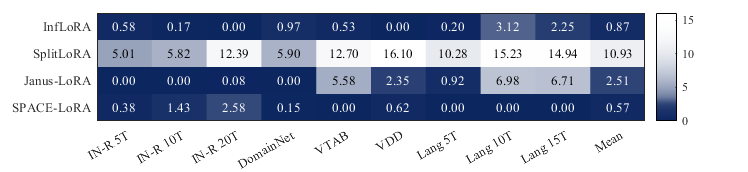}
    \caption{Gap to the best method on each benchmark (pp). Darker indicate a smaller gap.}
    \label{fig:regret}
\end{figure}

\paragraph{Results on Language Continual Learning.}
Table~\ref{tab:language_cl} reports results on the 5-, 10-,
and 15-task language benchmarks.
SPACE-LoRA achieves the highest final accuracy in the
5- and 10-task settings, reaching 71.88\% and 73.95\%,
respectively, with no forgetting on average in the 5-task
setting and 0.98 points of forgetting in the 10-task setting.
In the 15-task setting, SPACE-LoRA achieves 69.79\%
accuracy, exceeding InfLoRA by 2.25 points, with lower forgetting (2.53 versus 3.91 points).
Although SPACE-LoRA does not achieve the lowest forgetting
in any setting, it maintains competitive final accuracy
across all three task sequences.
These results complement the heterogeneous-task evaluation
on VTAB and VDD with continual adaptation across distinct
text classification tasks.
Figure~\ref{fig:regret} shows that SPACE-LoRA stays close to the best method on every benchmark, whereas the strongest baseline changes from benchmark to benchmark.

\subsection{Effect of Protection Coverage}
\label{sec:coverage_sweep}

The Fisher coverage threshold $\rho$ controls how much
historical information is covered by the protected
activation subspace.
We examine $\rho \in \{0.75, 0.80, 0.90, 0.95, 0.99\}$
on the 5-task ImageNet-R benchmark while keeping
the remaining training configuration fixed.
In addition to final accuracy and forgetting, we report
the average protected dimension $\bar{k}$.
Here, $\bar{k}$ averages the protected dimensions across
modules and incremental tasks, excluding the first task.

\begin{table}[t]
\centering
\caption{
Effect of Fisher coverage on 5-task ImageNet-R.
}
\label{tab:coverage_sweep}
\small
\setlength{\tabcolsep}{7pt}
\begin{tabular}{cccc}
\toprule
$\rho$
& Acc. $\uparrow$
& F $\downarrow$
& $\bar{k}$ \\
\midrule
0.75 & $78.12 \pm 0.26$ & 7.41 & 90 \\
0.80 & $78.35 \pm 0.37$ & 7.34 & 113 \\
0.90 & $\mathbf{79.09 \pm 0.31}$ & 4.82 & 211 \\
0.95 & $78.98 \pm 0.33$ & 4.90 & 322 \\
0.99 & $78.30 \pm 0.34$ & \textbf{3.50} & 517 \\
\bottomrule
\end{tabular}
\end{table}

Table~\ref{tab:coverage_sweep} illustrates the trade-off
between retaining previous knowledge and preserving
capacity for subsequent adaptation.
As $\rho$ increases from 0.75 to 0.99, the average
protected dimension grows from 90 to 517,
while forgetting decreases overall from 7.41 to
3.50 percentage points.
Final accuracy does not improve monotonically:
it increases to 79.09\% at $\rho=0.90$ and then
decreases to 78.30\% at $\rho=0.99$.
Thus, stronger protection can improve retention
without producing a corresponding improvement
in final predictive performance.

\subsection{Ablation on Protection Allocation}
\label{sec:allocation_ablation}

We next examine the role of Fisher information in
determining module-wise protection.
We compare Fisher-guided protection with uniform
allocation on 5-task ImageNet-R using ViT-B/16
and on the 10-task language benchmark using T5-large.
The Fisher-guided configuration uses $\rho=0.90$.
This ablation complements the coverage sweep by
examining how the protection rule affects final
accuracy and forgetting.

\begin{table}[t]
\centering
\caption{
Protection rule ablation on 5-task ImageNet-R and
10-task language continual learning with T5-large,
with $\rho=0.90$ for Fisher-based protection.
}
\label{tab:allocation_ablation}
\small
\setlength{\tabcolsep}{5pt}
\begin{tabular}{lcccc}
\toprule
& \multicolumn{2}{c}{ImageNet-R: 5 Tasks}
& \multicolumn{2}{c}{Language: 10 Tasks (T5-large)} \\
\cmidrule(lr){2-3}\cmidrule(lr){4-5}
Protection rule
& Acc. $\uparrow$ & F $\downarrow$
& Acc. $\uparrow$ & F $\downarrow$ \\
\midrule
Fisher (SPACE-LoRA)
& $\mathbf{79.09 \pm 0.31}$ & 4.82
& \textbf{73.95} & 0.98 \\
Uniform
& 76.97 & \textbf{4.40}
& 72.61 & \textbf{-0.12} \\
\bottomrule
\end{tabular}
\end{table}

Table~\ref{tab:allocation_ablation} shows that
Fisher-guided protection achieves higher final accuracy
than uniform allocation in both settings.
On ImageNet-R, it achieves 79.09\% accuracy compared
with 76.97\% for uniform allocation, an improvement
of 2.12 percentage points, while exhibiting slightly
higher forgetting (4.82 versus 4.40 points).
On the language benchmark, Fisher-guided protection
achieves 73.95\% accuracy compared with 72.61\%
for uniform allocation, an improvement of
1.34 percentage points.
These results support the effectiveness of
Fisher-guided protection in improving final
predictive performance over uniform allocation,
while showing that these gains do not arise
from lower forgetting.

\section{Conclusion}
In this work, we introduced SPACE-LoRA, a continual adaptation method that addresses catastrophic forgetting by suppressing the residual response of newly introduced LoRA branches. SPACE-LoRA projects the input factor of each new branch onto the orthogonal complement of selected historical activation directions, which guarantees zero residual response on the protected subspace while leaving the remaining directions for new task adaptation. A module-wise Fisher-coverage criterion further sets the protection size from the historical loss sensitivity along activation-ordered directions.
Experiments on vision and language benchmarks show that SPACE-LoRA reduces forgetting while remaining competitive, and the coverage analysis shows the trade-off between protecting previous tasks and learning new ones. Although Fisher-coverage sizing does not consistently outperform uniform sizing, protecting important historical activation directions is effective for continual low-rank adaptation.

\section*{AI Use Statement}
We used generative AI tools to assist with drafting and polishing
the manuscript. All AI-assisted text was reviewed and revised
by the authors, who take full responsibility for the final content.

\newpage

\bibliography{conference}
\bibliographystyle{conference}

\appendix
\section{Formal Properties of the Hard Protection Constraint}
\label{app:formal_properties}
This section provides a formal characterization of the hard input-space constraint used in SPACE-LoRA.
We consider a linear module \(m\) whose newly introduced LoRA branch at task \(t\) is given by
\begin{equation}
\Delta h_{m}^{(t)}(x)
=
\eta B_{m,t}A_{m,t}x,
\qquad
\eta = \frac{\alpha}{r},
\end{equation}
where
\begin{equation}
A_{m,t}\in\mathbb{R}^{r\times d_m},
\qquad
B_{m,t}\in\mathbb{R}^{o_m\times r}.
\end{equation}
Let
\begin{equation}
V_m
=
[v_{m,1},\ldots,v_{m,k_m}]
\in
\mathbb{R}^{d_m\times k_m},
\qquad
V_m^\top V_m = I,
\end{equation}
denote the protected basis selected for module \(m\), and let
\begin{equation}
P_m = V_mV_m^\top
\end{equation}
be the corresponding orthogonal projector.
After each optimizer update, the input factor is projected onto the orthogonal complement of the protected subspace as
\begin{equation}
A_{m,t}
\leftarrow
A_{m,t}(I-P_m).
\label{eq:app_projection}
\end{equation}

\subsection{Exact Nullification on the Protected Subspace}
\label{app:exact_nullification}

The projection in Eq.~\eqref{eq:app_projection} directly removes every component of \(A_{m,t}\) that acts on the protected directions.
Since \(P_m=V_mV_m^\top\) and \(V_m^\top V_m=I\),
\begin{align}
A_{m,t}(I-P_m)V_m
&=
A_{m,t}
\left(
V_m-V_mV_m^\top V_m
\right)
\\
&=
A_{m,t}(V_m-V_m)
=
0.
\end{align}
Therefore, the projected input factor satisfies
\begin{equation}
A_{m,t}V_m=0.
\label{eq:app_AV_zero}
\end{equation}
This property directly determines the response of the new residual branch to inputs contained in the protected subspace.
For any
\begin{equation}
x\in\operatorname{span}(V_m),
\end{equation}
there exists \(c\in\mathbb{R}^{k_m}\) such that \(x=V_mc\).
The corresponding residual response is then
\begin{align}
\Delta h_m^{(t)}(x)
&=
\eta B_{m,t}A_{m,t}x
\\
&=
\eta B_{m,t}A_{m,t}V_mc
\\
&=
0.
\label{eq:app_exact_nullification}
\end{align}
Thus, the new LoRA branch produces exactly zero response for inputs lying in the selected protected subspace, irrespective of the value of \(B_{m,t}\).

This property follows directly from the parameter projection rather than from a soft regularization objective.
It does not require a penalty term, a first-order approximation of the loss, or an assumption that the optimization gradient is approximately orthogonal to the protected directions.
As long as Eq.~\eqref{eq:app_AV_zero} is maintained, the new branch cannot introduce an additional response to an input that lies entirely in \(\operatorname{span}(V_m)\).

The scope of this result is local to the corresponding module and its input.
It guarantees that the newly added branch produces no additional response when the module receives an input within the protected subspace.
It does not require that every old task activation lie exactly in that subspace, nor does it imply that downstream activations remain unchanged after adaptation at earlier modules.
Therefore, the module-level property should not by itself be interpreted as a guarantee of zero forgetting at the network level.

\subsection{Interference Outside the Protected Subspace}
\label{app:interference_bound}

For an arbitrary input \(x\in\mathbb{R}^{d_m}\), the protected and unprotected components can be separated as
\begin{equation}
x
=
P_mx+(I-P_m)x.
\label{eq:app_decomposition}
\end{equation}
Because Eq.~\eqref{eq:app_AV_zero} also gives
\begin{equation}
A_{m,t}P_m
=
A_{m,t}V_mV_m^\top
=
0,
\end{equation}
the new adapter responds only to the component outside the protected subspace,
\begin{equation}
A_{m,t}x
=
A_{m,t}(I-P_m)x.
\label{eq:app_residual_input}
\end{equation}
Consequently,
\begin{align}
\left\|
\Delta h_m^{(t)}(x)
\right\|_2
&=
|\eta|
\left\|
B_{m,t}A_{m,t}(I-P_m)x
\right\|_2
\\
&\le
|\eta|
\left\|B_{m,t}\right\|_2
\left\|A_{m,t}\right\|_2
\left\|(I-P_m)x\right\|_2,
\label{eq:app_interference_bound}
\end{align}
where the inequality follows from submultiplicativity of the spectral norm.
For fixed adapter norms, the magnitude of the module-output perturbation is therefore controlled by how much of the input remains outside the protected subspace.
Inputs contained entirely in the protected subspace give zero residual response, while inputs close to that subspace can affect the new branch only through their remaining orthogonal component.

The size of this residual component can be related to the activation statistics used to construct \(V_m\).
Let the accumulated historical activation Gram matrix be
\begin{equation}
H_m
=
\sum_i x_i x_i^\top
=
U_m\Lambda_mU_m^\top,
\end{equation}
with eigenvalues ordered as
\begin{equation}
\lambda_{m,1}
\ge
\lambda_{m,2}
\ge
\cdots
\ge
\lambda_{m,d_m}
\ge
0.
\end{equation}
When
\begin{equation}
V_m=[u_{m,1},\ldots,u_{m,k_m}],
\end{equation}
the total historical activation energy remaining outside the protected subspace is
\begin{align}
\sum_i
\left\|
(I-P_m)x_i
\right\|_2^2
&=
\operatorname{tr}
\left[
(I-P_m)H_m
\right]
\\
&=
\sum_{j=k_m+1}^{d_m}
\lambda_{m,j}.
\label{eq:app_tail_energy}
\end{align}
Therefore, for a fixed protection dimension \(k_m\), selecting the leading eigenvectors of the historical activation Gram matrix minimizes the total squared residual energy of past activations outside the protected subspace.
The activation statistics thus determine the ordering of directions to be protected, while the Fisher-based allocation described in the main text determines how many of these directions are retained for each module.

\Eqref{eq:app_interference_bound} characterizes the residual response at the module level rather than task-level forgetting.
Its magnitude also depends on the learned norms of \(A_{m,t}\) and \(B_{m,t}\), which may vary across tasks and protection configurations.
In addition, an output perturbation introduced at an earlier module can alter the input received by subsequent modules.
For this reason, a small residual activation energy outside the protected subspace does not directly translate into an equivalent bound on the final network output.

\section{Construction of Historical Statistics}
\label{app:historical_statistics}

SPACE-LoRA uses two historical statistics to construct and evaluate the protected input directions: an activation Gram matrix and an input-space empirical Fisher matrix.
Both statistics are collected at the endpoint of each task and accumulated across previously observed tasks.
During later tasks, the protection subspace and its allocation are determined from these cached statistics without revisiting the raw training examples of previous tasks.

\subsection{Historical Activation Statistics}
\label{app:activation_statistics}
For each module \(m\), we summarize the input activations observed at the endpoint of task \(\tau\) using the uncentered Gram matrix
\begin{equation}
H_m^{(\tau)} = \sum_{i\in D_\tau} \sum_q x_{m,i,q}x_{m,i,q}^{\top},
\end{equation}
where \(x_{m,i,q}\in\mathbb{R}^{d_m}\) denotes the input activation at the \(q\)-th valid input position of example \(i\).
The historical activation statistic available before task \(t\) is
\begin{equation}
H_m^{<t} = \sum_{\tau<t} H_m^{(\tau)}.
\end{equation}
Hence, after completing task \(t\), the cached activation statistic can be updated incrementally as
\begin{equation}
H_m^{<t+1}=H_m^{<t}+H_m^{(t)},    
\end{equation}
without retaining activations or training examples from earlier tasks.

Then, We compute its eigendecomposition
\begin{equation}
H_m^{<t} = U_m\Lambda_mU_m^\top,
\end{equation}
where the eigenvalues are ordered in descending order.
The corresponding eigenvectors
\begin{equation}
U_m=[u_{m,1},\ldots,u_{m,d_m}]    
\end{equation}
define the ordered candidate directions for protection.

Importantly, the activation statistic determines both the directions and their prefix order.
SPACE-LoRA does not reorder these directions according to the Fisher sensitivity.
The Fisher is instead used only to determine how far this activation-ordered prefix should be extended for each module.
Given a protection dimension \(k_m\), the protected basis is therefore
\begin{equation}
V_m = [u_{m,1},\ldots,u_{m,k_m}].
\end{equation}

\subsection{Historical Fisher Sensitivity}
\label{app:fisher_statistics}
While the activation Gram determines the candidate protection directions, SPACE-LoRA uses an input-space empirical Fisher statistic to evaluate how sensitive each direction is to previously learned tasks.

For an example \(i\) from task \(\tau\), let
\begin{equation}
G_{m,i}^{(\tau)} = \nabla_{W_m}\ell_i^{(\tau)} \in \mathbb{R}^{o_m\times d_m}
\end{equation}
denote the gradient of the supervised loss with respect to the effective weight of module \(m\).
At the endpoint of task \(\tau\), we compute this statistic using the current task classifier and the observed ground-truth labels.
We define the task-level input-space empirical Fisher as
\begin{equation}
F_m^{(\tau)} = \frac{1}{N_\tau} \sum_{i=1}^{N_\tau} \left(G_{m,i}^{(\tau)}\right)^\top G_{m,i}^{(\tau)}.
\end{equation}
This matrix corresponds to the input-dimensional partial trace of the observed-label empirical Fisher associated with the effective weight \(W_m\).
The historical Fisher available before task \(t\) is accumulated using the number of examples in each previous task:
\begin{equation}
F_{m,\mathrm{past}}^{(t)} = \frac{\sum_{\tau<t} N_\tau F_m^{(\tau)}}{\sum_{\tau<t} N_\tau }.
\end{equation}
In practice, this quantity can be maintained using the accumulated Fisher numerator together with the total number of examples, without storing task-specific training data.

For the \(j\)-th activation direction \(u_{m,j}\), we define its historical sensitivity as
\begin{equation}
p_{m,j}^{(t)} = u_{m,j}^{\top} F_{m,\mathrm{past}}^{(t)} u_{m,j}.
\end{equation}
Equivalently, this quantity can be written as the average squared sensitivity along \(u_{m,j}\) over all examples from previously observed tasks:
\begin{equation}
p_{m,j}^{(t)} = \frac{1}{\sum_{\tau<t}N_\tau} \sum_{\tau<t} \sum_{i=1}^{N_\tau} \left\| G_{m,i}^{(\tau)}u_{m,j} \right\|_2^2.
\end{equation}
Since \(F_{m,\mathrm{past}}^{(t)}\) measures squared gradient responses along input directions, each directional sensitivity \(p_{m,j}^{(t)}\) is nonnegative.
Moreover, because \(U_m=[u_{m,1},\ldots,u_{m,d_m}]\) forms a complete orthonormal basis, the sensitivities over all activation directions sum to the total historical sensitivity:
\begin{align}
\sum_{j=1}^{d_m}p_{m,j}^{(t)}
&= \sum_{j=1}^{d_m} u_{m,j}^{\top} F_{m,\mathrm{past}}^{(t)} u_{m,j} \\
&= \operatorname{tr} \left( U_m^\top F_{m,\mathrm{past}}^{(t)} U_m \right) \\
&= \operatorname{tr} \left( F_{m,\mathrm{past}}^{(t)} \right).
\end{align}
Therefore, the cumulative sensitivity of the first \(k\) activation directions, \(\sum_{j=1}^{k}p_{m,j}^{(t)}\), represents the amount of historical sensitivity captured by that activation-ordered prefix.
This identity provides the basis for the Fisher-coverage ratio used in Section~\ref{sec:fisher}.

\section{Additional Experimental Details}
\label{sec:appendix_exp}

\subsection{Vision Experimental Setup}
\label{sec:appendix_vision}

For ImageNet-R, the 5, 10, and 20 task settings use the same underlying image split and class ordering. This allows us to examine the effect of increasing the number of sequential learning stages while keeping the overall class set fixed.

For DomainNet, we use five domains consisting of clipart, infograph, painting, real, and sketch, while excluding Quickdraw. The 345 classes are divided into five disjoint tasks with 69 classes per task. Images from all five retained domains are included in every task.

For VTAB, we use a class incremental benchmark consisting of five datasets, Resisc45, DTD, Oxford IIIT Pets, EuroSAT, and Oxford Flowers102, in this order. Each dataset contributes 10 classes and constitutes a separate task. This results in five tasks and 50 classes in total. We use the officially released data and hold out 10\% of the training images from each class as a validation set. This split yields 1,615 training images, 181 validation images, and 8,619 test images.

For VDD, we use nine of the ten datasets in the Visual Domain Decathlon and exclude ImageNet. The nine datasets are Aircraft, CIFAR 100, Daimler Pedestrian Classification, DTD, GTSRB, Omniglot, SVHN, UCF101, and VGG Flowers. Each dataset constitutes one task in this order, such that every task introduces both a new domain and a new label set. The resulting benchmark contains 2,128 classes in total, with the number of classes per task ranging from 2 to 1,623. Since the labels of the official test split are not publicly available, we use the official validation split as the test set and hold out 10\% of the training images from each class for validation. This results in 156,598 training images, 17,178 validation images, and 64,439 test images.

\subsection{Language Experimental Setup}
\label{sec:appendix_language}

We evaluate continual language adaptation using a sequence of 15 text classification and natural language understanding datasets. Table~\ref{tab:datasets} presents the datasets in the order in which they are introduced to the model. The 5 task setting uses the first five datasets, the 10 task setting uses the first ten datasets, and the 15 task setting uses the complete sequence. Therefore, each longer setting extends the preceding sequence without changing the order of previously introduced tasks.

\begin{table}[t]
\centering
\caption{
Datasets used in the language continual learning experiments in their training order.
}
\label{tab:datasets}
\small
\setlength{\tabcolsep}{5pt}
\begin{tabular}{clll}
\toprule
Step & Dataset & Source Suite & Task Type \\
\midrule
1  & AG News & Standard CL & Topic Classification \\
2  & Amazon  & Standard CL & Sentiment Analysis \\
3  & Yelp    & Standard CL & Sentiment Analysis \\
4  & DBpedia & Standard CL & Topic Classification \\
5  & Yahoo   & Standard CL & Question Answering \\
\midrule
6  & MNLI   & GLUE & Natural Language Inference \\
7  & QQP    & GLUE & Paraphrase Detection \\
8  & RTE    & GLUE & Natural Language Inference \\
9  & SST-2  & GLUE & Sentiment Analysis \\
10 & IMDB   & Independent & Sentiment Analysis \\
\midrule
11 & WiC     & SuperGLUE & Word Sense Disambiguation \\
12 & CB      & SuperGLUE & Natural Language Inference \\
13 & COPA    & SuperGLUE & Question Answering \\
14 & MultiRC & SuperGLUE & Question Answering \\
15 & BoolQ   & SuperGLUE & Boolean Question Answering \\
\bottomrule
\end{tabular}
\end{table}

For most tasks, we construct a fixed dataset containing 1,000 training samples, 200 validation samples, and up to 500 test samples. The validation and test sets do not overlap whenever sufficient labeled data are available.

For GLUE and SuperGLUE datasets whose public test labels are unavailable, we construct evaluation splits from the corresponding labeled validation data. The resulting test sets for RTE and WiC contain 77 and 438 samples, respectively. For smaller datasets such as CB and COPA, we use the available labeled evaluation samples for performance measurement.

All experiments with T5 large use AdamW with a constant learning rate, a batch size of 8, and 3 epochs per task. We use LoRA with rank 8 and scaling parameter 16 for the query and value projections in all attention modules. A new adapter is introduced for each task, and the maximum input length is 128 tokens.

We select the learning rate once using the five task setting with seed 0 and use the selected value for all methods and sequence lengths. No method specific learning rate tuning is performed. All compared methods use the same data, optimizer, and training schedule, while their method specific hyperparameters follow the default settings of their official implementations.

\subsection{Evaluation Protocol}
\label{sec:appendix_metrics}

The model is evaluated after completing each task in the sequence. We report average accuracy after the final task to measure overall predictive performance and backward transfer to measure changes in performance on previously learned tasks. All methods are evaluated using the same task ordering, dataset splits, training schedule, and evaluation protocol.

\begin{algorithm}[t]
\caption{Continual Adaptation with SPACE-LoRA}
\label{alg:space_lora}
\begin{algorithmic}[1]
\Require Task datasets $\{D_t\}_{t=1}^{T}$, protected modules $\mathcal{M}$, LoRA rank $r$, coverage target $\rho$
\State Initialize $H_m^{<1}\gets 0$ and $F_{m,\mathrm{past}}^{(1)}\gets 0$ for all $m\in\mathcal{M}$
\State Initialize the number of previous-task examples $N_{<1}\gets 0$

\For{$t=1,\ldots,T$}

    \State Freeze the base model and all LoRA adapters learned from previous tasks
    \State Introduce a new rank-$r$ LoRA branch $(A_{m,t},B_{m,t})$ for each $m\in\mathcal{M}$

    \For{each module $m\in\mathcal{M}$}

        \If{$t=1$}
            \State $k_m\gets 0$, $P_m\gets 0$
        \Else
            \State Compute
            $ H_m^{<t} = U_m\Lambda_mU_m^\top $
            with eigenvectors ordered by decreasing eigenvalue

            \For{$j=1,\ldots,d_m$}
                \State
                $ p_{m,j}^{(t)} \gets u_{m,j}^{\top} F_{m,\mathrm{past}}^{(t)} u_{m,j} $
            \EndFor

            \If{$\operatorname{tr}(F_{m,\mathrm{past}}^{(t)})=0$}
                \State $k_m\gets 0$
            \Else
                \State Find the smallest
                $ k\in\{0,\ldots,d_m-r\} $
                satisfying
                \[ \frac{\sum_{j=1}^{k}p_{m,j}^{(t)}}{ \operatorname{tr} \left( F_{m,\mathrm{past}}^{(t)} \right) } \ge \rho \]
                \If{such $k$ exists}
                    \State $k_m\gets k$
                \Else
                    \State $k_m\gets d_m-r$
                \EndIf
            \EndIf

            \State
            $ V_m \gets [u_{m,1},\ldots,u_{m,k_m}] $
            \State
            $ P_m \gets V_mV_m^\top $
        \EndIf
    \EndFor

    \For{each optimization step on $D_t$}
        \State Compute the current-task loss and update the trainable parameters
        \For{each module $m\in\mathcal{M}$}
            \State Project the input factor
            $ A_{m,t} \gets A_{m,t}(I-P_m) $
        \EndFor
    \EndFor

    \For{each module $m\in\mathcal{M}$}
        \State Collect endpoint activations on $D_t$ and compute
        \[ H_m^{(t)} \gets \sum_{i\in D_t} \sum_q x_{m,i,q}x_{m,i,q}^{\top} \]

        \State Using the current task classifier and ground-truth labels, compute
        $ G_{m,i}^{(t)} \gets \nabla_{W_m}\ell_i^{(t)} $
        for each $i\in D_t$

        \State Compute
        \[ F_m^{(t)} \gets \frac{1}{N_t} \sum_{i=1}^{N_t} \left(G_{m,i}^{(t)}\right)^\top G_{m,i}^{(t)} \]

        \State Update the historical activation statistic
        $ H_m^{<t+1} \gets H_m^{<t}+H_m^{(t)} $

        \State Update the historical Fisher
        \[ F_{m,\mathrm{past}}^{(t+1)} \gets \frac{ N_{<t}F_{m,\mathrm{past}}^{(t)} + N_tF_m^{(t)}}{ N_{<t}+N_t } \]
    \EndFor
    \State
    $ N_{<t+1} \gets N_{<t}+N_t $
\EndFor
\end{algorithmic}
\end{algorithm}

\end{document}

%% file: math_commands.tex
\usepackage{amsmath,amsfonts,bm}

\def\eqref#1{equation~\ref{#1}}
\def\Eqref#1{Equation~\ref{#1}}

\def\1{\bm{1}}

\DeclareMathAlphabet{\mathsfit}{\encodingdefault}{\sfdefault}{m}{sl}
\SetMathAlphabet{\mathsfit}{bold}{\encodingdefault}{\sfdefault}{bx}{n}